\documentclass[10pt,twocolumn,letterpaper]{article}

 \usepackage{cvpr}              
\definecolor{cvprblue}{rgb}{0.21,0.49,0.74}
\usepackage[pagebackref,breaklinks,colorlinks,allcolors=cvprblue]{hyperref}

\usepackage[utf8]{inputenc} 
\usepackage[T1]{fontenc}    
\usepackage{url}            
\usepackage{booktabs}       
\usepackage{amsfonts}       
\usepackage{nicefrac}       
\usepackage{microtype}      
\usepackage{xcolor}         
\usepackage{amssymb}
\usepackage{pifont}
\usepackage{amsmath}
\usepackage{footnote}
\usepackage{xspace}
\usepackage{wrapfig}
\usepackage{tabularx}
\usepackage{multirow}
\usepackage{amssymb}
\usepackage{bbm}
\usepackage{graphicx}
\usepackage{enumitem}
\usepackage{wrapfig}
\usepackage{tablefootnote}
\usepackage{dsfont}
\usepackage{refcount}
\makeatletter
\@namedef{ver@eso-pic.sty}{2020/10/14 v3.0a eso-pic (CVPR shim)}
\makeatother
\usepackage{pdfpages}

\DeclareMathOperator*{\argmax}{arg\,max}

\def\paperID{15009} 
\def\confName{CVPR}
\def\confYear{2026}

\title{Exploring Hierarchical Consistency and Unbiased Objectness\\ for Open-Vocabulary Object Detection}

\author{
	Sanghoon Lee$^{1}$\quad Geon Lee$^{1}$\quad Hyekang Park$^{1}$\quad Bumsub Ham$^{1,2}$\thanks{Corresponding author}\\[5pt]
	$^{1}$Yonsei University\quad $^{2}$Korea Institute of Science and Technology (KIST) \\
	{\url{https://cvlab.yonsei.ac.kr/projects/HCC}}
}

\begin{document}
\maketitle
\begin{abstract}

    Conventional object detectors typically operate under a closed-set assumption, limiting recognition to a predefined set of base classes seen during training. Open-vocabulary object detection (OVD) addresses this limitation by leveraging vision-language models (VLMs) to generate pseudo labels for novel object classes. However, existing OVD methods suffer from two critical drawbacks: (1) inaccurate class label assignments, as VLMs are optimized for image-level predictions rather than the region-level predictions required for pseudo labeling, and (2) and unreliable objectness scores from region proposal networks (RPNs) trained exclusively on base object classes. To address these issues, we propose a novel pseudo labeling framework for OVD. Our approach introduces a hierarchical confidence calibration (HCC) technique, which ensures reliable class label estimation by assessing consistency across hierarchical semantic levels (class, super- and sub-category). We also present LoCLIP, a parameter-efficient adaptation of CLIP that incorporates an objectness token to mitigate base class bias problem of RPNs and provide reliable objectness estimations for novel object classes. Extensive experiments on standard OVD benchmarks, including COCO and LVIS, demonstrate that our approach clearly sets a new state of the art, validating the effectiveness of our approach. 
\end{abstract}    
\section{Introduction}\label{sec:intro}


Conventional object detectors~\cite{carion2020end, tan2020efficientdet, ren2016faster, liu2016ssd, redmon2016you} typically assume a closed-set scenario, and they are restricted to recognizing only a fixed set of base object classes provided at training time. The closed-set assumption limits scalability in real-world environments, where arbitrary object classes appear~\cite{bansal2018zero, joseph2021towards}. To overcome this limitation, open-vocabulary object detection (OVD)~\cite{zareian2021open, gu2021open} has been introduced to recognize novel object classes, while exploiting annotations for a set of base object classes only.

Recent works for OVD employ vision-language models~(VLMs)~\cite{radford2021learning, jia2021scaling} that provide semantically aligned image-text representations~\cite{zhang2024vision}. In particular, VLMs are typically used to generate pseudo labels for objects of novel classes~\cite{wang2023object, zhao2024taming, wang2024marvelovd, zhao2022exploiting}. To this end, given a candidate region of an object~(\ie, an object proposal) obtained from an RPN~\cite{ren2016faster}, OVD methods assign pseudo labels to candidate regions that are likely to contain objects of novel classes. The object proposals, however, often correspond to irrelevant regions~(\eg,~misaligned or background regions) that do not contain objects of novel classes. In order to filter out the irrelevant regions, current methods compute confidence scores from classification probabilities obtained using a VLM (\eg, CLIP~\cite{radford2021learning}), where the text features representing novel classes are used as classifiers. The object proposals with insufficient confidence scores are regarded as irrelevant regions, and they are discarded during the pseudo labeling process. To further consider the localization accuracy of candidate regions, objectness scores from the RPN are typically used. Note that the RPN is trained on a set of base class objects, and current OVD methods assume that the RPN is able to provide reliable objectness scores even for unseen object classes during training. Accordingly, pseudo labels are selectively assigned to object proposals that show sufficiently high confidence scores from CLIP and objectness scores from the RPN. 

Current OVD approaches to exploiting pseudo labels~\cite{wang2023object, zhao2024taming, wang2024marvelovd, zhao2022exploiting} have two major drawbacks. First, confidence scores obtained from CLIP does not well-represent the presence of objects of novel classes. Consequently, the pseudo labels are dominated by irrelevant regions. On COCO~\cite{lin2014microsoft}, as an example, $76\%$ of the generated pseudo labels from CLIP correspond to background regions~\cite{wang2024marvelovd}. This mainly stems from the fact that CLIP is optimized for image-level predictions, while pseudo labeling in OVD requires making region-level predictions~\cite{lin2022learning}. Second, the RPN is exclusively trained using annotations for objects of base classes. That is, all other classes, including novel ones, are treated as background during training. This implies that the RPN could offer unreliable objectness scores for objects of novel classes~\cite{joseph2021towards}. This is more pronounced in the context of pseudo labeling, where the RPN is expected to identify novel object instances as foreground, despite having learned to treat them as background during its training.

\begin{figure}
    \centering
    \includegraphics[width=1.0\columnwidth]{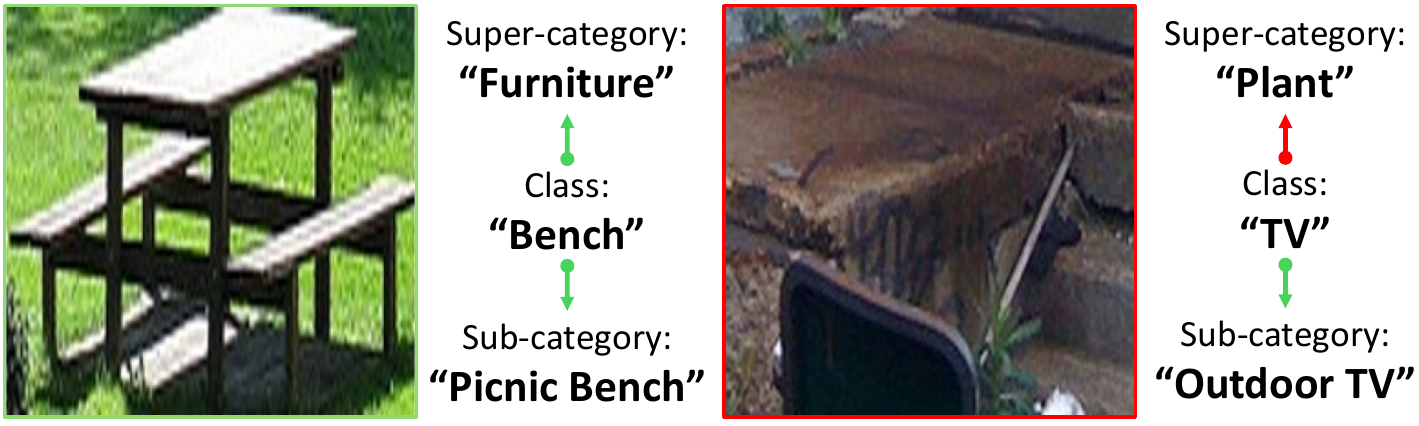}
    \vspace{-0.65cm}
    \caption{Visualization of hierarchical consistency of candidate regions. We use CLIP~\cite{radford2021learning} to classify regions at class, super- and sub-category levels based on LLM-generated hierarchy on COCO~\cite{lin2014microsoft}. We can see that an accurately localized region (left) yields hierarchically consistent predictions (indicated by green arrows), whereas a background region (right) yields inconsistent predictions (indicated by red arrows). Experimental details and quantitative analysis are in the supplementary materials.}\label{fig:teaser}
    \vspace{-0.4cm}
\end{figure}


In this paper, we present a novel pseudo labeling framework for OVD that addresses the aforementioned drawbacks of current approaches in estimating reliable confidence and objectness scores of candidate regions. To this end, we propose a hierarchical confidence calibration (HCC) technique that adjusts confidence scores of CLIP, by incorporating super- and sub-categories of novel object classes. Specifically, we observe that regions that well-localize an object tend to yield hierarchically consistent predictions, while irrelevant regions do not~(Fig.~\ref{fig:teaser}). Based on this observation, we present a confidence calibration technique that promotes confidence scores of candidate regions whose classification predictions at different hierarchy levels are consistent, while suppressing the confidence scores for the opposite case. We also introduce an objectness estimation method, dubbed LoCLIP, using parameter-efficient adaptation of CLIP. LoCLIP introduces an additional objectness token into CLIP that quantifies how well a candidate region localizes an object. Compared to the RPN, objectness estimations from LoCLIP are less biased towards base object classes, offering more reliable objectness estimations for novel object classes. Experimental results on standard OVD benchmarks, including COCO~\cite{lin2014microsoft} and LVIS~\cite{gupta2019lvis}, demonstrate the effectiveness of our approach, which clearly sets a new state of the art. The main contributions can be summarized as follows:

\begin{itemize}[leftmargin=*]
\item We present a novel pseudo labeling framework for OVD that calibrates confidence scores of candidate regions by exploiting super- and sub-categories of novel object classes.

\item We introduce LoCLIP that estimates the objectness of candidate regions through a parameter-efficient adaptation scheme. The LoCLIP addresses the base class bias of RPNs, while being computationally efficient. 

\item We achieve a new state of the art on standard benchmarks for OVD including COCO~\cite{lin2014microsoft} and LVIS~\cite{gupta2019lvis}, and demonstrate the effectiveness of our approach through extensive experiments with ablation studies. 
\end{itemize}

\section{Related work}

We describe in this section recent works pertinent to ours, including OVD and VLMs using a language hierarchy. \\

\begin{figure*}[t]
    \centering
    \includegraphics[width=0.85\textwidth]{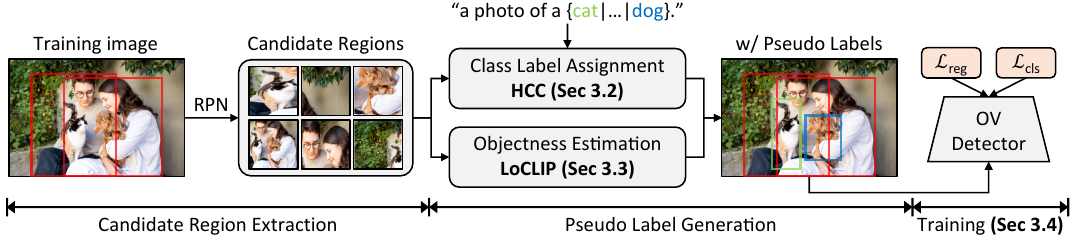}
    \vspace{-0.3cm}
    \caption{Overview of our framework for OVD, which mainly consists of three steps. First, a set of candidate regions is extracted from an image using an RPN. For each candidate region, we employ the HCC technique for selectively assigning a class label, while LoCLIP estimates an objectness score. Pseudo labels are assigned only to regions with hierarchically consistent predictions and sufficiently high objectness scores. These pseudo labels, together with ground-truth annotations for base object classes, are then used to train an OV detector with classification and regression losses. See the main text for details.}\label{fig:overview}
    \vspace{-0.5cm}
\end{figure*}

\noindent\textbf{OVD.}~~OVD aims at learning semantically aligned region-text representations, such that a detector can localize and classify novel object classes in a zero-shot manner. To this end, many works attempt to train VLMs that are able to make region-level predictions~\cite{zhong2022regionclip, yao2022detclip, minderer2022simple, cheng2024yolo, kim2023region, kim2023contrastive}. For example, RegionCLIP~\cite{zhong2022regionclip} aligns region and text representations using lots of image-caption pairs~\cite{sharma2018conceptual}, where the regions are extracted using an RPN~\cite{ren2016faster}. The VLMs optimized for region-level predictions provide competitive performance on standard benchmarks for OVD. However, training VLMs tailored for object detection typically involves large-scale pretraining using a lot of image-text pairs~\cite{schuhmann2022laion, jia2021scaling, kebe2021a}, which is computationally demanding. 

To overcome this challenge, OVD methods instead exploit VLMs trained using image-level contrastive learning~\cite{radford2021learning, jia2021scaling}. These methods can be categorized into two groups: The first line of works~\cite{gu2021open, ma2022open, du2022learning, zang2022open, wang2023object, shi2023edadet, wu2023clipself} focuses on distilling knowledge~\cite{hinton2015distilling} from VLMs to OV detectors. They employ knowledge distillation techniques to encourage region-level features, extracted from the detector, to imitate visual features from the VLM. However, OVD methods using knowledge distillation techniques are typically biased toward base object classes~\cite{shi2023edadet}, and offer suboptimal performance for novel ones. Many distillation-based approaches thus ensemble classification probabilities for base and novel classes separately at inference time. This increases the computational cost for inference, while requiring tedious hyper-parameter search for better results~\cite{gu2021open, kuo2022f}. Another line of works attempt to generate pseudo labels for novel object classes~\cite{wang2023object, zhao2024taming, wang2024marvelovd, zhao2022exploiting}. In particular, they typically exploit the classification confidence and objectness scores, which are measured using a VLM and an RPN, respectively, as criteria for selecting candidate proposals to assign pseudo labels. For example, the work of~\cite{zhao2022exploiting} computes the geometric mean of the classification confidence and the objectness score for each object proposal, and assigns pseudo labels only for the proposal, whose classification confidence and objectness score exceed a certain threshold. The generated pseudo labels, together with ground-truth annotations of base classes, are then used to train an OV detector. In the pseudo labeling regime, effectively determining whether a candidate region contains a novel object class is crucial, as candidate regions mostly correspond to irrelevant background regions~\cite{wang2024marvelovd}. To handle distractive noise in the generated pseudo labels, recent works~\cite{zhao2024taming, wang2024marvelovd} further employ self-training techniques~\cite{tang2017multiple, xu2021end}. For example, the work of~\cite{zhao2024taming} uses separate prediction branches for base and novel classes to prevent noisy pseudo labels from distracting learning base object classes. Rather than employing dedicated strategies to handle noisy labels, our work focuses on establishing reliable pseudo labels, which can then be used directly to train an OV detector.

To further enhance the recognition capabilities of OV detectors, many works~\cite{bangalath2022bridging, gao2022open, huang2024open, li2024cliff, zhou2022detecting, lin2022learning, ma2023codet} exploit additional supervision from auxiliary datasets. Such datasets provide images with class labels~\cite{ridnik2021imagenet} or captions~\cite{sharma2018conceptual, chen2015microsoft}, offering strong priors about the presence of objects in images. For example, Detic~\cite{zhou2022detecting} leverages a subset of ImageNet-21K~\cite{ridnik2021imagenet} whose class labels overlap with LVIS~\cite{gupta2019lvis} to compensate for the lack of supervision for novel object classes, thereby improving overall detection performance. While OVD methods that rely on auxiliary datasets generally provide better results, acquiring these additional supervision sources requires significant labeling efforts. In contrast, our framework does not rely on any auxiliary datasets for training.  \\

\noindent\textbf{VLMs with a language hierarchy.}~~Recent studies~\cite{novack2023chils, ge2023improving, liu2024shine, huang2024open} have shown that leveraging hierarchical relationships among object classes can significantly enhance the prediction performance of VLMs. By incorporating semantic hierarchy structure of objects, these approaches enable more informed reasoning about visual categories, enhancing generalization and improving robustness across various tasks. For image classification on ImageNet~\cite{deng2009imagenet}, recent works~\cite{novack2023chils, ge2023improving} propose to re-rank initial predictions of CLIP using the hierarchical structure of WordNet~\cite{miller1995wordnet}. They show that the hierarchy information enhances the recognition capabilities of an off-the-shelf VLM. Building on this idea, OVD methods also adopt the language hierarchy to improve detection performance, by augmenting classifiers of OV detectors~\cite{liu2024shine} or class labels themselves~\cite{huang2024open}, using super-/sub-categories of target object classes. Different from prior works, we leverage hierarchical structure between object classes for determining whether or not a local image region contains an object of interest. Moreover, while current methods that exploit a hierarchy structure focus primarily on the classification subtask, we additionally consider the localization accuracy of local image regions for OVD.


\section{Approach}

We present in this section an overview of our framework~(Sec.~\ref{sec:overview}), which mainly consists of class label assignment using HCC~(Sec.~\ref{sec3.2:HCC}) and objectness estimation using LoCLIP~(Sec.~\ref{sec3.3:LoCLIP}), and describe an overall training process~(Sec.~\ref{sec3.4:training}).

\subsection{Overview}\label{sec:overview}

Following the standard protocol~\cite{gu2021open}, we split object classes into two disjoint sets of base and novel classes, denoted by $C_{B}$ and $C_{N}$, respectively, and assume that the class names of $C_{N}$ are given at training time, following other OVD methods~\cite{wang2023object, zhao2024taming, wang2024marvelovd, zhao2022exploiting}. We train an OV detector that can recognize objects of both base and novel classes in $C_{B} \cup C_{N}$, with ground-truth bounding boxes and object labels for base classes in $C_{B}$ only. To this end, we generate pseudo labels for objects of novel classes in $C_{N}$, facilitating the OV detector to recognize novel object classes, thereby allowing it to generalize beyond base classes provided for training.

We provide in Fig.~\ref{fig:overview} an overview of our framework for generating pseudo labels of novel object classes, which mainly consists of three steps. We first employ an RPN~\cite{ren2016faster} to generate candidate regions for the pseudo label assignment. Notably, while the candidate regions indeed contain objects of novel classes, most of them correspond to irrelevant ones, \eg, misaligned or background regions~\cite{wang2024marvelovd}. To address this problem, we present a HCC technique and LoCLIP for assigning pseudo labels only to candidate regions that localize objects of novel classes well. The HCC technique considers the consistency of VLM’s predictions across multiple hierarchy levels (class, super- and sub-categories) to estimate a confidence level of a VLM for its prediction of candidate regions. Meanwhile, LoCLIP estimates objectness of candidate regions that quantify how well a region localizes an object of any class. The pseudo labels are only assigned to candidate regions that yield sufficient confidence and objectness scores obtained using HCC and LoCLIP, respectively. Once the pseudo labels are established, the OV detector is trained with both the ground-truth annotations for base classes and the pseudo labels for novel classes, using classification and regression losses~\cite{ren2016faster, he2017mask}.

\subsection{Class label assignment}\label{sec3.2:HCC}
We specify each novel object class into a textual form using, \eg, ``a photo of a [class].''. Each sentence is fed into the VLM to obtain a text feature representing a novel object class, denoted by $\mathbf{c}_{n} \in \mathbb{R}^{d}$, where $n=1,\dots,|C_{N}|$ and $d$ is the feature dimension. We then compute similarities between visual features of candidate regions and text features of novel object classes. Concretely, we denote an image by $\mathbf{x}$ and bounding box coordinates of a candidate region by $b \in \mathbb{R}^{4}$. We extract a visual feature $\mathbf{v}_{b} \in \mathbb{R}^{d}$ for the region $b$ using an image encoder $\mathcal{V}$ of a VLM as follows:
\begin{equation}
	\mathbf{v}_{b} = \mathcal{V}(\mathbf{x}_{b}),
\end{equation}
where~$\mathbf{x}_{b} = \text{crop}(\mathbf{x}, b)$ and the $\text{crop}(\cdot,\cdot)$ operation extracts a corresponding region of an image w.r.t bounding box coordinates. We then compute classification probabilities $\mathbf{p}\in\mathbb{R}^{|C_{N}|}$ of a candidate region for novel classes as follows:
\begin{equation}\label{eq:confidence}
	\mathbf{p}({n}) = \frac{\exp(s_{n})}{\sum_{j} \exp(s_{j})},~~~~\text{where}~~~~
    s_{n} = 
    \frac{\mathbf{c}_{n} \mathbf{v}_{b}^{\top}}
    { \| \mathbf{c}_{n} \| \, \| \mathbf{v}_{b} \| },
\end{equation}
and $\Vert\cdot\Vert$ measures L2-norm of a vector. Note that we omit the temperature parameter for brevity.
We can define a confidence score~$\hat{p}$ for the region~$b$ as the maximum probability across novel object classes,~\ie, $\hat{p}=\max(\mathbf{p})$. It measures a confidence level of a VLM for its classification for a candidate region, \ie, how likely a region contains an object of novel class. Current OVD methods~\cite{wang2023object, zhao2024taming, wang2024marvelovd, zhao2022exploiting} assign a class label selectively only when the confidence score $\hat{p}$ is sufficiently high. This is, however, problematic, since VLMs are not fit to represent local image regions~\cite{lin2022learning}. In particular, these VLMs do not discriminate between well-localized and irrelevant image regions. This suggests that the irrelevant regions could have high confidence scores, and thus applying a threshold for the confidence~$\hat{p}$ alone might not filter out incorrect class labels assignments effectively. 

In light of this, we observe that a region that tightly covers an object tends to yield hierarchically consistent predictions across class, super- and sub-category levels, whereas irrelevant regions do not~(Fig.~\ref{fig:teaser}). This observation aligns with recent findings for hierarchical image classification~\cite{park2024visually}, which demonstrate that CLIP~\cite{radford2021learning} provides hierarchically consistent predictions for object-centric images, but the consistency degrades in complex scenes containing multiple object classes. This suggests that incorporating hierarchical consistency could offer valuable cues for discarding irrelevant regions during the pseudo labeling process, which is a crucial aspect for suppressing incorrect class label assignments to irrelevant regions~\cite{wang2024marvelovd}. \\

\begin{figure}[t]
    \centering
    \includegraphics[width=1.0\columnwidth]{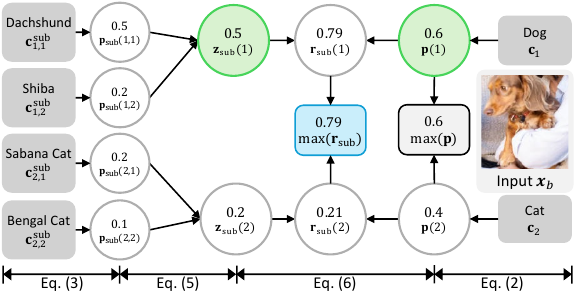}
    \vspace{-0.5cm}
    \captionsetup{font=small}
    \caption{An illustration of the operations in the HCC technique for an input image (right), where novel object classes are defined as \textit{dog} and \textit{cat}, and $K$ is set to $2$. Here, the class label estimation from the class level and the sub-category level are consistent for \textit{dog}, represented in green nodes. Consequently, the HCC technique boosts the confidence score from $\max(\mathbf{p})=0.6$ to $\max(\mathbf{r}_{\operatorname{sub}})=0.79$. See text for details.}\label{fig:hcc}
\end{figure}

\noindent\textbf{HCC.}~~We propose a HCC technique that adjusts confidence scores, $\hat{p}$, based on hierarchical consistency across predictions from class, super- and sub-category levels. That is, we increase confidence scores for candidate regions that yield hierarchically consistent predictions, while suppressing the confidence scores in the opposite case. To this end, we obtain super- and sub-categories of object classes by prompting a LLM, \eg, GPT-OSS~\cite{agarwal2025gpt}, as done in~\cite{novack2023chils, liu2024shine}. We query the LLM to provide multiple super-/sub-categories for each novel object class, in order to consider diverse hierarchical relationships. Specifically, we obtain a hierarchy for novel object classes, where each class is associated with $K$ super- and sub-categories. We denote a text feature for the $k$-th super- and sub-category of the $n$-th novel class by $\mathbf{c}^{\sup}_{n, k} \in \mathbb{R}^{d}$ and $\mathbf{c}^{\operatorname{sub}}_{n, k} \in \mathbb{R}^{d}$, respectively, where $k=1,\dots,K$. The HCC technique adjusts confidence scores, $\hat{p}$, by leveraging super- and sub-categories separately. In the following, we describe the HCC process using the established sub-categories.

We first compute classification probabilities at the sub-category level, denoted by $\mathbf{p}_{\operatorname{sub}}\in\mathbb{R}^{|C_{N}| \times K}$, as follows:
\begin{equation}\label{eq:subcategory}
    \mathbf{p}_{\operatorname{sub}}(n,k) = \frac{\exp(s^{\operatorname{sub}}_{n,k})}{\sum_{m}\sum_{l} \exp(s^{\operatorname{sub}}_{m,l})},
\end{equation}
where
\begin{equation}
    s^{\operatorname{sub}}_{n,k} = 
    \frac{\mathbf{c}^{\operatorname{sub}}_{n,k} \mathbf{v}_{b}^{\top}}
    { \| \mathbf{c}^{\operatorname{sub}}_{n,k} \| \, \| \mathbf{v}_{b} \| }.
\end{equation}
Note that the softmax in Eq.~\eqref{eq:subcategory} is computed across all $|C_{N}| \times K$ sub-categories. The resulting matrix $\mathbf{p}_{\operatorname{sub}}$ therefore represents a single probability distribution over all sub-categories (\ie, $\sum_{n,k} \mathbf{p}_{\operatorname{sub}}(n,k) = 1$). We then aggregate the $K$ sub-category scores for each class $n$ to make $\mathbf{p}_{\operatorname{sub}}$ and $\mathbf{p}$ dimensionally compatible. Specifically, we use max pooling to consider the most relevant sub-category within each class, yielding class-wise sub-category scores, $\mathbf{z}_{\operatorname{sub}} \in \mathbb{R}^{|C_{N}|}$, as follows:
\begin{equation}\label{eq:qsub}
	\mathbf{z}_{\operatorname{sub}}(n) = \max(\mathbf{p}_{\operatorname{sub}}(n)),
\end{equation}
where $\mathbf{p}_{\operatorname{sub}}(n) \in \mathbb{R}^{K}$ is the $n$-th row vector of $\mathbf{p}_{\operatorname{sub}}$. We use the scores $\mathbf{z}_{\operatorname{sub}}$ to reweight the class probabilities $\mathbf{p}$, thereby ensembling the predictions from the class and sub-category levels, to obtain the calibrated probabilities $\mathbf{r}_{\operatorname{sub}} \in \mathbb{R}^{|C_{N}|}$ as follows:
\begin{equation}\label{eq:rsub}
	\mathbf{r}_{\operatorname{sub}}({n}) = \frac{\mathbf{p}(n)\mathbf{z}_{\operatorname{sub}}(n)}
	{\sum_{m}\mathbf{p}(m)\mathbf{z}_{\operatorname{sub}}(m)}.
\end{equation}
The denominator in Eq.~\eqref{eq:rsub} ensures that the elements in $\mathbf{r}_{\operatorname{sub}}$ sum to $1$. With the calibrated scores $\mathbf{r}_{\operatorname{sub}}$, we can define adjusted confidence score as the maximum score across novel object classes, \ie, $\max(\mathbf{r}_{\operatorname{sub}})$. In particular, when predictions at the class and sub-category levels are consistent, HCC guarantees that the adjusted confidence score w.r.t sub-categories is greater than or equal to the uncalibrated one. That is,
\begin{equation}\label{eq:property1}
	\argmax(\mathbf{p}) = \argmax(\mathbf{z}_{\operatorname{sub}})~~\Rightarrow~~\max(\mathbf{r}_{\operatorname{sub}}) \ge \hat{p}.
\end{equation}
Proof for Eq.~\eqref{eq:property1} can be found in the supplementary materials. We also show in the supplementary materials that we can enforce the opposite case to hold when the predictions are inconsistent as follows:
\begin{equation}\label{eq:property2}
	\argmax(\mathbf{p}) \ne \argmax(\mathbf{z}_{\operatorname{sub}})
	~~\Rightarrow~~\max(\mathbf{r}_{\operatorname{sub}}) < \hat{p}
\end{equation}
with a minor effort. The calibrated probabilities obtained using super-categories, denoted by $\mathbf{r}_{\operatorname{sup}} \in \mathbb{R}^{|C_{N}|}$, are similarly defined. We provide in Fig.~\ref{fig:hcc} an example case for the operations in HCC.

We then combine the calibrated scores from both super- and sub-categories to obtain a final confidence score, $\hat{r}$, as follows:
\begin{equation}\label{eq:calibration}
\hat{r}=\max(\mathbf{r}),~~\text{where}~~\mathbf{r}(n) = \frac{\mathbf{r}_{\operatorname{sub}}(n) + \mathbf{r}_{\operatorname{sup}}(n)}{2}.
\end{equation}
We selectively assign pseudo labels to candidate regions that yield sufficient adjusted confidence scores, $\hat{r}$. That is, we assign a pseudo class label $\hat{y}_{b}$ for the region $b$ as follows:
\begin{equation}\label{eq:assignment}
\hat{y}_{b} = \argmax(\mathbf{r})~~\text{if}~~\hat{r} \ge \gamma,
\end{equation}
where $\gamma$ is a predefined threshold. Compared to using $\hat{p}$ as a confidence score, using the adjusted score $\hat{r}$ allows to address the core issue in current methods~\cite{wang2023object, zhao2024taming, wang2024marvelovd, zhao2022exploiting}, where the pseudo labels mostly correspond to background regions~\cite{wang2024marvelovd}. \\

\subsection{Objectness estimation}\label{sec3.3:LoCLIP}

Current methods exploit objectness scores from the RPN to estimate the objectness of candidate regions. However, the RPN is trained on a set of annotations for base object classes, where novel ones are considered as background. This suggests that the RPN is likely to provide incorrect objectness scores for novel object classes. This is more problematic in the context of pseudo label generation, where the RPN is expected to identify objects of novel classes as foreground, even though it is trained to suppress those same instances as background during training.

\begin{figure}[t]
    \centering
    \includegraphics[width=0.9\columnwidth]{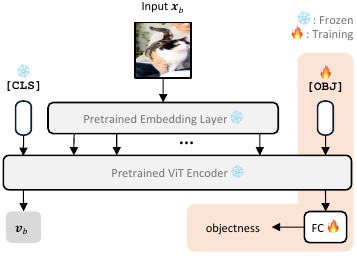}
    \vspace{-0.3cm}
    \captionsetup{font=small}
    \caption{Illustration of LoCLIP. LoCLIP appends a learnable \texttt{[OBJ]} token into a pretrained ViT encoder of CLIP. The output feature for the \texttt{[OBJ]} token is passed through a FC layer, and predicts an objectness score of an input patch~$\mathbf{x}_{b}$. Throughout the adaptation process, all pretrained components, including the \texttt{[CLS]} token, embedding layers and ViT encoders, are kept frozen.}\label{fig:loclip}
    \vspace{-0.5cm}
\end{figure}


\noindent\textbf{LoCLIP.}~~To overcome this problem, we present a localization-aware CLIP, dubbed LoCLIP~(Fig.~\ref{fig:loclip}), which is a parameter-efficient adaptation of a CLIP model, to estimate objectness of candidate regions. To this end, LoCLIP introduces a learnable $\texttt{[OBJ]}$ token, and appends it to the image encoder of CLIP, enabling the token to interact with frozen CLIP features. The output feature for the $\texttt{[OBJ]}$ token is then fed into a fully-connected~(FC) layer with a sigmoid activation to predict objectness scores for candidate regions. LoCLIP provides better objectness scores with less bias toward base object classes, compared to the RPN, for the following reasons: First, the $\texttt{[OBJ]}$ token in LoCLIP interacts with frozen CLIP features that are not exclusively biased toward base object classes. Second, LoCLIP involves significantly less number of parameters exclusively tuned for the base object classes. For example, the number of parameters in LoCLIP, including ones in the FC layer, is about $3$K, whereas the RPN used in the works of~\cite{wang2023object, zhao2024taming, wang2024marvelovd, zhao2022exploiting} uses $41$M parameters. This suggests that our LoCLIP is likely to be less overfitted towards base object classes while generalizing well to novel ones. We consider candidate regions whose objectness scores from the LoCLIP is lower than a threshold $\tau$ as background regions, and discard them in the pseudo labeling process.

To implement LoCLIP, we adopt a masked attention technique~\cite{lee2023read}, that enables retaining visual features from the image encoder of CLIP. In this way, a single forward pass through the LoCLIP model provides visual features and objectness scores of candidate regions simultaneously, facilitating an efficient pseudo labeling pipeline. Note that we could exploit other parameter-efficient strategies,~\eg, by employing an additional FC layer~\cite{houlsby2019parameter} that takes visual features of CLIP as inputs for estimating the objectness. This strategy indeed provides better results compared to the RPN (See Sec.~\ref{sec:discussion}). However, its prediction does not directly involve patch-level features that encode useful local information for estimating the objectness. On the contrary, LoCLIP exploits the \texttt{[OBJ]} token that directly interacts with local features through self-attention mechanism in vision transformers~\cite{dosovitskiy2020image}~(ViTs). It is thus more effective in estimating objectness scores for objects belonging to novel classes. To train the \texttt{[OBJ]} token and an additional FC layer, we use binary cross entropy loss. Specifically, we assign a binary label for each candidate region, depending on its intersection over union~(IoU) with any objects belonging to the base classes. Note that LoCLIP converges using only $1\%$ of the entire training images from COCO~\cite{lin2014microsoft} and LVIS~\cite{gupta2019lvis}, and takes about $5$ minutes to train on a single NVIDIA A6000 GPU. We refer to the supplementary material for more details on the training process of LoCLIP.

\subsection{Training}\label{sec3.4:training}

We train an OV detector with pseudo labels for novel object classes and ground-truth annotations for base classes. In particular, each pseudo label includes an objectness score from LoCLIP and a confidence $\hat{r}$ from HCC, which indicate how well the region is localized and how confidently it is classified, respectively. For example, a higher objectness score suggests that the region tightly covers an object of a novel class. Similarly, a higher confidence indicates that a VLM is confident in assigning a class label for a given region. Based on these characteristics, we re-weight classification and regression losses for pseudo labels, with confidence and objectness scores, respectively. Concretely, we denote the bounding box coordinates of the $i$-th prediction from an OV detector as $\hat{u}_{i} \in \mathbb{R}^{4}$ and its predicted class label as $\hat{c}_{i}$. Similarly, we denote by $u_{i}$ and $c_{i}$ bounding box coordinates and a class label, respectively, obtained from ground-truth annotations for base classes or pseudo labels for novel classes. Note that when the class label belongs to a novel object class, \ie, $c_{i} \in C_{N}$, it is associated with a calibrated confidence score from HCC and an objectness score from LoCLIP, which we denote by $z_{i}$ and $o_{i}$, respectively. Our loss $\mathcal{L}_{i}$ for the $i$-th prediction can be represented as follows:
\begin{equation}
	\mathcal{L}_{i} = \begin{cases}
		\mathcal{L}_{\mathrm{cls}}(\hat{c}_{i}, c_{i}) + \mathds{1}_{[c_{i} \in C_{B}]} \mathcal{L}_{\mathrm{reg}}(\hat{u}_{i}, u_{i}) & \text{if}~c_{i} \in C_{B} \cup \{ bg \} \\
		z_{i} \mathcal{L}_{\mathrm{cls}}(\hat{c}_{i}, c_{i}) + o_{i}\mathcal{L}_{\mathrm{reg}}(\hat{u}_{i}, u_{i}) & \text{if}~c_{i} \in C_{N},
	\end{cases}
\end{equation}
where $\mathcal{L}_{\mathrm{cls}}$ and $\mathcal{L}_{\mathrm{reg}}$ are classification and bounding box regression losses, respectively, and $\mathds{1}$ is an indicator function that outputs $1$ if an argument is true and $0$ otherwise. Our overall loss is defined as $\mathcal{L} = \sum_{i}\mathcal{L}_{i}$. In this way, supervisory signals from pseudo labels with low objectness or confidence scores are suppressed during training. This enables the OV detector to focus on more reliable regions, thereby mitigating the distractive influence of misaligned or uncertain pseudo labels and improving its overall detection performance.
\section{Experiments}

In this section, we describe implementation details~(Sec.~\ref{sec4.1:implementation}), and compare our method with the state of the art~(Sec.~\ref{sec:sota}). We then provide in-depth analysis of our components, including HCC and LoCLIP~(Sec.~\ref{sec:discussion}). More results, including qualitative results, analyses on different LLMs, and hyper-parameters, together with additional discussions can be found in the supplementary material.

\subsection{Implementation details}\label{sec4.1:implementation}

\noindent\textbf{Dataset.}~~We train and evaluate our model under OV settings using the COCO~\cite{lin2014microsoft} and LVIS~\cite{gupta2019lvis} datasets, referred to as OV-COCO and OV-LVIS, respectively. For OV-COCO, we split the $65$ object classes into $48$ base and $17$ novel classes. The training set contains $118,287$ images, while the evaluation set consists of $4,836$ images\footnote{Following the standard protocol~\cite{zhao2022exploiting, wang2024marvelovd}, we exclude images without any novel object classes from the COCO validation split.}. We report box mean AP$_{50}$ for base and novel classes separately, denoted by AP$_{50}^{B}$ and AP$_{50}^{N}$, respectively. For OV-LVIS, which contains $1,203$ object classes, we set the $337$ rare classes as novel ones, and the remaining $866$ object classes as base classes. We report mask mAP for novel and all object classes separately, denoted by AP$^{N}_{m}$ and AP$^{All}_{m}$, respectively. 

\noindent\textbf{Network.}~~We adopt Faster R-CNN~\cite{ren2016faster} with a ResNet-50~\cite{he2016deep} backbone for OV-COCO, and Mask R-CNN~\cite{he2017mask} with a ResNet-50 backbone for OV-LVIS, together with a class-agnostic mask prediction head. Following~\cite{wu2023aligning, wang2023object}, both detection networks are initialized with SoCo~\cite{wei2021aligning} pretrained weights. For the classifier, we set the weights using text features derived from the prompt ``a photo of a [class]'', and set the background class weight to zero, as done in~\cite{zhao2022exploiting, wang2024marvelovd, kuo2022f}. For the VLM, we employ CLIP~\cite{radford2021learning} with a ViT-B/32~\cite{dosovitskiy2020image} image encoder, with the official weights released by OpenAI. For the LLM, we primarily use GPT-OSS-120b~\cite{agarwal2025gpt}. 

\noindent\textbf{Training.}~~On OV-COCO, we train our network using the $1\times$ scheduling scheme, which corresponds to $90$k training iterations. The learning rate is set to $0.02$ and decreased by a factor of $10$ at the $60$k-th and $80$k-th iterations with a total batch size of $16$. By default, we use random flip only for data augmentation during training. For OV-LVIS, following the protocol described in~\cite{wu2023aligning, wang2023object}, we use the $2\times$ schedule and train our model for $180$k iterations, where the learning rate is set to $0.02$ and decreased by a factor of $10$ at the $120$k-th and $160$k-th iterations. For both datasets, we use a warmup strategy that gradually increases a learning rate from $0$ to $0.02$ during the first $1$k training iterations. Training the detectors take approximately $6$ hours for OV-COCO and $16$ hours for OV-LVIS, with $8$ NVIDIA A6000 GPUs. To further demonstrate the generalization capability of our approach, we also apply our method on top of CLIPSelf~\cite{wu2023clipself}. In this case, we replace base annotation used in the original work with our generated pseudo labels and train a detector using the official implementation provided by the authors. To generate pseudo labels, we set $\gamma=0.8$ and $\tau=0.3$ for OV-COCO and $\gamma=0.6$ and $\tau=0.2$ for OV-LVIS. We establish an initial LLM-generated hierarchy by using $K=10$ for super-categories and $K=30$ for sub-categories on both datasets. Note that we further refine the initial super-/sub-categories through an LLM-driven process, which we describe in detail in the supplementary materials. 

\subsection{Results}\label{sec:sota}
We provide in Table~\ref{tab:comparison-coco} quantitative comparisons of our method with the state of the art on OV-COCO~\cite{lin2014microsoft}. For a fair comparison, all listed method use RCNN-style detectors, and do not use additional datasets for training. From the table, we can see that our model clearly sets a new state of the art in terms of AP$_{50}^{N}$. This demonstrates the effectiveness of our approach in establishing reliable pseudo labels for novel object classes. Our approach even outperforms SAS-Det~\cite{zhao2024taming}, which exploits RegionCLIP~\cite{zhong2022regionclip}, a VLM tailored for detection, compared to CLIP in ours. Compared to MarvelOVD~\cite{wang2024marvelovd}, which requires additional stages to handle distractive noise in pseudo labels, our approach provides better performance with less computational cost for training. Specifically, the total training time for MarvelOVD is $1.5\times$ longer than ours, despite using the same $1\times$ scheduling scheme. This result highlights that obtaining reliable pseudo labels, rather than adopting computationally intensive self-training strategies, is more significant for OVD. Table~\ref{tab:comparison-lvis} compares the performance of various OVD methods on the OV-LVIS~\cite{gupta2019lvis} dataset, further demonstrating the effectiveness of our approach. It shows that our method achieves the highest performance in terms of AP$_{m}^{N}$. Furthermore, we can see that our generated pseudo labels also improve the performance of CLIPSelf~\cite{wu2023clipself} on both OV-COCO and OV-LVIS datasets. This demonstrates that our framework can generalize well to various backbones and training strategies. We provide in the supplementary material a more comprehensive comparison with state-of-the-art methods, including ones leveraging additional datasets.

\begin{table}[t]
\centering
\captionsetup{font=small}
\centering
\caption{Comparison with state-of-the-art methods on the OV-COCO dataset~\cite{lin2014microsoft}. To ensure a fair comparison, all listed methods employ RCNN-style detectors~\cite{ren2016faster, he2017mask}. We report the mean and standard deviation across $3$ independent runs. ${\dagger}$ indicates results obtained from our re-implementation.}\vspace{-0.3cm}
\label{tab:comparison-coco}
\small
\begin{tabular*}{\columnwidth}{@{\extracolsep{\fill}} l c c c c}
\toprule
Method & \begin{tabular}{@{}c@{}}Backbone\\Network\end{tabular} & AP$_{50}^{N}$ & \textcolor{gray}{AP$_{50}^{B}$} \\
\midrule
ViLD~\cite{gu2021open} & RN50 & 27.6 & \textcolor{gray}{59.5} \\
F-VLM~\cite{kuo2022f} & RN50 & 28.0 & \textcolor{gray}{40.2} \\
OADP\tablefootnote{The results for OADP~\cite{wang2023object}, reported in the original publication, were obtained using LSJ data augmentation~\cite{ghiasi2021simple}, which is inconsistent with the experimental setting described in the paper. We thus report the results without the LSJ technique, obtained from the \href{https://github.com/LutingWang/OADP/tree/d641d28c77e551f748019f81ee4574b2a63a6366}{official repository}.\label{fn:oadp}}~\cite{wang2023object} & RN50 & 31.3 & \textcolor{gray}{-} \\
VL-PLM~\cite{zhao2022exploiting} & RN50 & 32.3 & \textcolor{gray}{54.0} \\
RALF~\cite{kim2024retrieval} & RN50 & 33.4 & \textcolor{gray}{54.5} \\
BARON~\cite{wu2023aligning} & RN50 & 34.0 & \textcolor{gray}{60.4} \\
MarvelOVD$^{\dagger}$~\cite{wang2024marvelovd} & RN50 & 35.4 & \textcolor{gray}{56.5} \\
SAS-Det~\cite{zhao2024taming} & RN50 & \underline{37.4} & \textcolor{gray}{58.5} \\
Ours & RN50 & \textbf{38.9}\tiny{$\pm$0.3} & \textcolor{gray}{59.5\tiny{$\pm$0.2}} \\
\midrule
CLIPSelf$^{\dagger}$~\cite{wu2023clipself} & ViT-L/14 & \underline{41.3} & \textcolor{gray}{65.5} \\
Ours & ViT-L/14 & \textbf{44.0}\tiny{$\pm$0.2} & \textcolor{gray}{65.8\tiny{$\pm$0.1}} \\
\bottomrule
\end{tabular*}
\vspace{-0.2cm}
\end{table}

\begin{table}[t]
\centering
\caption{Comparison with state-of-the-art methods on the OV-LVIS dataset~\cite{gupta2019lvis}. To ensure a fair comparison, all listed methods employ RCNN-style detectors~\cite{he2017mask}. We report the mean and standard deviation across $3$ independent runs. ${\dagger}$ indicates results obtained from our re-implementation.}\vspace{-0.3cm}
\label{tab:comparison-lvis}
\small
\begin{tabular*}{\columnwidth}{@{\extracolsep{\fill}} l c c c}
\toprule
Method & \begin{tabular}{@{}c@{}}Backbone\\Network\end{tabular} & AP$_{m}^{N}$ & \textcolor{gray}{AP$_{m}^{All}$} \\
\midrule
ViLD~\cite{gu2021open} & RN50 & 16.6 & \textcolor{gray}{25.5} \\
F-VLM~\cite{kuo2022f} & RN50 & 18.6 & \textcolor{gray}{24.2} \\
BARON~\cite{wu2023aligning} & RN50 & 19.2 & \textcolor{gray}{26.5} \\
DetPro~\cite{du2022learning} & RN50 & 19.8 & \textcolor{gray}{25.9} \\
OADP\footref{fn:oadp}~\cite{wang2023object} & RN50 & 19.9 & \textcolor{gray}{-} \\
SAS-Det~\cite{zhao2024taming} & RN50 & \underline{20.9} & \textcolor{gray}{26.1} \\
Ours & RN50 & \textbf{21.7}\tiny{$\pm$0.4} & \textcolor{gray}{26.0\tiny{$\pm$0.2}} \\
\midrule
CLIPSelf$^{\dagger}$~\cite{wu2023clipself} & ViT-B/16 & \underline{25.1} & \textcolor{gray}{24.5} \\
Ours & ViT-B/16 & \textbf{25.5}\tiny{$\pm$0.2} & \textcolor{gray}{24.7\tiny{$\pm$0.1}} \\
\bottomrule
\end{tabular*}
\vspace{-0.5cm}
\end{table}

\subsection{Discussion}\label{sec:discussion}

\noindent\textbf{Ablation study.}~~We present in Table~\ref{tab:ablation} an ablation analysis of HCC and LoCLIP. We train object detectors with pseudo labels, generated by various combinations of our components, including super- and sub-categories of the language hierarchy for HCC, and LoCLIP. The baseline in \ding{172} corresponds to our implementation of VL-PLM~\cite{zhao2022exploiting}. From the table, we can make the following observations: (1)~As shown in \ding{173} and \ding{174}, exploiting either super- or sub-categories of object classes provides more reliable confidence estimations, improving performance drastically for novel object classes. In particular, leveraging sub-categories with $\mathbf{r}_{\operatorname{sub}}$ provides better pseudo labels compared to using super-categories with $\mathbf{r}_{\operatorname{sup}}$. We suspect this stems from the fact that CLIP generally performs better when given fine-grained textual descriptions (sub-categories) than coarse-grained ones (super-categories), possibly because fine-grained concepts are known to appear more frequently~\cite{xu2024benchmarking} in its pretraining dataset~\cite{schuhmann2022laion}. Nevertheless, using both super- and sub-categories in \ding{175} provides complementary performance gains. (2)~LoCLIP provides performance improvements in \ding{176}. This demonstrates the clear advantage of leveraging LoCLIP over the RPN used in \ding{172} for estimating objectness scores for novel object classes. (3)~Finally, incorporating all components, as in \ding{177}, achieves the best performance, suggesting that the proposed components are complementary to each other. We point out that both HCC and LoCLIP share the common objective of preventing irrelevant regions from being assigned pseudo labels. However, they achieve this goal differently. The HCC technique focuses on capturing class-specific cues, while LoCLIP focuses on class-agnostic properties. Note that obtaining reliable pseudo labels also boost the performance of base classes, as it suppresses noise in training OV detectors.

\newcolumntype{C}{>{\centering\arraybackslash}X}
\begin{table}
\captionsetup{font=small}
\caption{Ablative analysis using different combinations of our approach, including HCC and LoCLIP, on OV-COCO~\cite{lin2014microsoft}. $r^{\operatorname{sub}}_{n}$ and $r^{\operatorname{sup}}_{n}$ indicates the use of corresponding terms in Eq.~\eqref{eq:calibration}.}\vspace{-0.3cm}
\label{tab:ablation}
\small
\begin{tabularx}{\columnwidth}{C|CC|C|CC}
\toprule
& \multicolumn{2}{c|}{HCC} & \multirow{2}{*}{LoCLIP} & \multirow{2}{*}{AP$_{50}^{N}$} & \multirow{2}{*}{\textcolor{gray}{AP$_{50}^{B}$}} \\
& $\mathbf{r}_{\operatorname{sub}}$ & $\mathbf{r}_{\operatorname{sup}}$ & & & \\
\midrule
\ding{172} &  &  &  & 32.2 & \textcolor{gray}{58.3} \\
\midrule
\ding{173} & \ding{51} &  &  & 36.8 & \textcolor{gray}{59.3} \\
\ding{174} &  & \ding{51} &  & 36.0 & \textcolor{gray}{59.5} \\
\ding{175} & \ding{51} & \ding{51} &  & \underline{37.8} & \textcolor{gray}{59.4} \\
\midrule
\ding{176} &  &  & \ding{51} & 33.9 & \textcolor{gray}{59.0} \\
\midrule
\ding{177} & \ding{51} & \ding{51} & \ding{51} & \textbf{38.9} & \textcolor{gray}{59.5} \\
\bottomrule
\end{tabularx}
\vspace{-0.5cm}
\end{table}

\noindent\textbf{LoCLIP.} For object proposals generated by RPN, we compute the correlation coefficients between their IoU with ground-truth boxes and their objectness scores, where the scores are obtained from the RPN, Adapter~\cite{houlsby2019parameter}, and LoCLIP, for unseen object classes. Specifically, we divide object classes into odd and even splits, and use one split for training, while exploiting the other for evaluation. Note that a higher correlation coefficient indicates that the corresponding score reflects the objectness of candidate regions more accurately. We can see from Table~\ref{tab:iou-correlation} that the RPN shows low correlation scores, even close to zero, suggesting that it does not quantify how well proposals cover objects of unseen classes effectively. This also indicates that using RPN scores for objectness, as in previous works~\cite{wang2023object, zhao2024taming, wang2024marvelovd, zhao2022exploiting}, would be suboptimal. Adapter improves the correlation coefficients, but it exploits image-level features (\ie, the output feature for the \texttt{[CLS]} token) to estimate the objectness, providing slightly worse performance than LoCLIP. This can be attributed to LoCLIP using local features for predicting objectness, in contrast to Adapter. Overall, LoCLIP provides the best results, demonstrating its effectiveness in estimating reliable objectness scores for novel object classes. We provide qualitative comparisons in the supplementary materials. 

\noindent\textbf{Efficiency.} We measure runtime to generate pseudo labels and compare with VL-PLM~\cite{zhao2022exploiting}, which does not involve HCC and LoCLIP compared to our method. On average, the VL-PLM method takes $0.872$ seconds per image to generate a set of pseudo labels, while our method, including HCC and LoCLIP, takes $0.892$ seconds, with only a negligible $2.3\%$ increase in total. We attribute the efficiency of our method to the following reasons: (1) LoCLIP retains the visual features of the CLIP model, allowing it to be used for both estimating objectness and extracting visual features from candidate regions. In practice, the visual features from LoCLIP are directly used for class label assignments with the HCC technique. (2) HCC is simple by design, requiring only the matrix operations in Eqs.~\eqref{eq:subcategory}-\eqref{eq:rsub}. 

\newcolumntype{C}{>{\centering\arraybackslash}X}
\begin{table}[t]
\centering
\captionsetup{font=small}
\caption{Comparisons of correlation coefficients for RPN, Adapter~\cite{houlsby2019parameter} and LoCLIP~(ours) for unseen object classes on OV-COCO~\cite{lin2014microsoft}. SR:~Spearman's $\rho$. KT:~Kendall's $\tau$.}\vspace{-0.3cm}
\label{tab:iou-correlation}
\small
\begin{tabularx}{\columnwidth}{C|CCC}
  \toprule
  Metric & RPN & Adapter~\cite{houlsby2019parameter} & LoCLIP \\
  \midrule
  \multicolumn{4}{c}{Even $\rightarrow$ Odd} \\
  \midrule
  SR $(\uparrow)$ & 0.038 & 0.456 & \textbf{0.473} \\
  KT $(\uparrow)$ & 0.024 & 0.313 & \textbf{0.326} \\
  \midrule
  \multicolumn{4}{c}{Odd $\rightarrow$ Even} \\
  \midrule
  SR $(\uparrow)$ & 0.002 & 0.434 & \textbf{0.449} \\
  KT $(\uparrow)$ & 0.001 & 0.300 & \textbf{0.310} \\
  \bottomrule
\end{tabularx}
\vspace{-0.5cm}
\end{table}

\section{Conclusion}

We have presented a novel pseudo labeling framework for OVD that addresses the limitations of existing approaches. To this end, we have introduced HCC technique that exploits a language hierarchy structure of object classes, providing more reliable confidence scores of candidate regions. We have also proposed LoCLIP that employs parameter-efficient adaptation of CLIP, and offers unbiased objectness estimations compared to an RPN. Extensive experiments demonstrate the effectiveness of our approach, which clearly sets a new state of the art on standard OVD benchmarks with a negligible increase in overall computational cost.

\let\thefootnote\relax\footnotetext{\hspace*{-1.5em}\textbf{Acknowledgements.} This work was partly supported by IITP grant funded by the Korea government~(MSIT)~(No.~RS-2022-00143524, Development of Fundamental Technology and Integrated Solution for Next-Generation Automatic Artificial Intelligence System and No.~2022-0-00124, RS-2022-II220124, Development of Artificial Intelligence Technology for Self-Improving Competency-Aware Learning Capabilities) and the KIST Institutional Program (Project No.2E33001-24-086).}

\clearpage
{
    \small
    \bibliographystyle{ieeenat_fullname}
    \bibliography{main}
}

\clearpage


\section*{Supplementary material}

# Exploring Hierarchical Consistency and Unbiased Objectness for Open-Vocabulary Object Detection

## Supplementary Material

In this supplementary material, we provide more details (Sec. S1), discussions (Sec. S2), results (Sec. S3) and limitations and future work (Sec. S4).

## S1. More details

In this section, we provided detailed descriptions for our experimental setup for obtaining results in Fig. 1 (Sec. S1.1), and proof/analysis for Eqs. (7) and (8) (Sec. S1.2). We then describe the training details for LoCLIP (Sec. S1.3), and a process for refining a LLM-generated hierarchy (Sec. S1.4).

### S1.1. Experimental setup for Fig. 1

Our approach is based on the observation that regions that well-localize an object tend to yield hierarchically consistent predictions, while irrelevant ones (*e.g.*, misaligned and background regions) do not. To demonstrate this observation, we leverage the class names and the ground-truth bounding box annotations of the base object classes in OV-COCO [19]. Specifically, we first establish a semantic hierarchy (*i.e.*, super- and sub-categories) for the base object classes using GPT-OSS-120b [1] as the LLM. For example, this process yields *picnic bench* as a sub-category of the class *bench*, which itself belongs to the super-category *furniture*. We then gather object proposals from an RPN [26] and categorize them based on their IoU with the ground-truth boxes. The proposals with an IoU score higher than $0.8$ are labeled as foreground (well-localized), while those with an IoU score lower than $0.5$ are labeled as background (irrelevant). We use this experimental setup to visualize the hierarchical consistency for well-localized and irrelevant regions separately in Fig. 1. Note that this analysis is performed using the class names and the ground-truth annotations for base object classes only. We provide in Fig. S2 more visualizations and quantitative analysis in Sec. S3.

### S1.2. Eqs. (7) and (8)

**Proof of Eq. (7).** We wish to show that

$$\arg\max(\mathbf{p}) = \arg\max(\mathbf{z}_{\text{sub}}) \;\Rightarrow\; \max(\mathbf{r}_{\text{sub}}) \geq \hat{p}, \tag{S1}$$

where $\hat{p} = \max(\mathbf{p})$. Let $l$ be the common index of the maximum elements, defined as:

$$l = \arg\max(\mathbf{p}) = \arg\max(\mathbf{z}_{\text{sub}}). \tag{S2}$$

Invoking the generalized mean inequality, specifically for orders set to 1 and $\infty$, we have:

$$\sum_m \mathbf{p}(m)\mathbf{z}_{\text{sub}}(m) \leq \max(\mathbf{z}_{\text{sub}}) = \mathbf{z}_{\text{sub}}(l). \tag{S3}$$

Applying Eq. (S3) to the definition of the $l$-th element of $\mathbf{r}_{\text{sub}}$, we obtain:

$$\begin{aligned}
\mathbf{r}_{\text{sub}}(l) &= \frac{\mathbf{p}(l)\mathbf{z}_{\text{sub}}(l)}{\sum_m \mathbf{p}(m)\mathbf{z}_{\text{sub}}(m)} \\
&\geq \frac{\mathbf{p}(l)\mathbf{z}_{\text{sub}}(l)}{\mathbf{z}_{\text{sub}}(l)} \\
&= \mathbf{p}(l).
\end{aligned} \tag{S4}$$

Here, Eq. (S4) holds because replacing the denominator $\sum_m \mathbf{p}(m)\mathbf{z}_{\text{sub}}(m)$ with its upper bound $\mathbf{z}_{\text{sub}}(l)$ yields a lower bound for the fraction. Finally, since $\mathbf{p}(l) = \hat{p}$, it follows that

$$\max(\mathbf{r}_{\text{sub}}) \geq \hat{p}. \tag{S5}$$

**Property in Eq. (8).** In the following, we demonstrate that Eq. (8) holds under two conditions: 1) $\max(\mathbf{p})$ is sufficiently large, *e.g.*, exceeding $0.5$, and 2) the score distribution of $\mathbf{z}_{\text{sub}}$ is sufficiently flat. We further show that these conditions can be enforced into our framework with minimal effort.

Let $l = \arg\max(\mathbf{p})$ denote the index of the maximum element in $\mathbf{p}$. We characterize the flatness of $\mathbf{z}_{\text{sub}}$ by bounding the ratio between its maximum and minimum values with a small constant $\beta \geq 1$ as follows:

$$1 \leq \frac{\max(\mathbf{z}_{\text{sub}})}{\min(\mathbf{z}_{\text{sub}})} \leq \beta. \tag{S6}$$

To prove that $\max(\mathbf{r}_{\text{sub}}) < \max(\mathbf{p})$, we analyze two cases. First, we examine the value of $\mathbf{r}_{\text{sub}}$ at the index $l$. Under the inconsistency assumption (*i.e.*, $\arg\max(\mathbf{z}_{\text{sub}}) \neq l$), the element $\mathbf{z}_{\text{sub}}(l)$ is strictly less than the maximum of $\mathbf{z}_{\text{sub}}$. Given the condition in Eq. (S6) and $\mathbf{z}_{\text{sub}}(l) < \max(\mathbf{z}_{\text{sub}})$, the mean $\sum_m \mathbf{p}(m)\mathbf{z}_{\text{sub}}(m)$ strictly exceeds the value $\mathbf{z}_{\text{sub}}(l)$. That is,

$$\mathbf{r}_{\text{sub}}(l) = \mathbf{p}(l)\left(\frac{\mathbf{z}_{\text{sub}}(l)}{\sum_m \mathbf{p}(m)\mathbf{z}_{\text{sub}}(m)}\right) < \mathbf{p}(l). \tag{S7}$$

Second, for any index $n \neq l$, we can represent $\mathbf{r}_{\text{sub}}(n)$ as follows:

$$\mathbf{r}_{\text{sub}}(n) = \mathbf{p}(n)\frac{\mathbf{z}_{\text{sub}}(n)}{\sum_m \mathbf{p}(m)\mathbf{z}_{\text{sub}}(m)}. \tag{S8}$$

Note that the scaling factor $\frac{\mathbf{z}_{\text{sub}}(n)}{\sum_m \mathbf{p}(m)\mathbf{z}_{\text{sub}}(m)}$ is strictly bounded by $\beta$, since $\sum_m \mathbf{p}(m)\mathbf{z}_{\text{sub}}(m) \geq \min(\mathbf{z}_{\text{sub}})$.

Given the condition that $\mathbf{p}(l)$ is sufficiently large, implying $\mathbf{p}(n)$ is small, it follows that

$$\mathbf{p}(n) \cdot \beta < \mathbf{p}(l). \tag{S9}$$

Since $\beta$ is close to 1, this condition holds naturally for any $\max(\mathbf{p}) > 0.5$, implying $\mathbf{r}_{\text{sub}}(n) < \max(\mathbf{p})$. Combining both cases, we can conclude that $\max(\mathbf{r}_{\text{sub}}) < \max(\mathbf{p})$.

In practice, we find that applying a minimum threshold to $\hat{p}$ is sufficient to ensure that Eq. (8) holds. This is because $\mathbf{z}_{\text{sub}}$ and $\mathbf{z}_{\text{sup}}$ are derived by max-pooling the probability distribution computed over the entire set of sub- and super-categories, respectively. Since the softmax operation is applied across a large number of sub-/super-categories, the score is inherently distributed, resulting in a flat distribution. Consequently, explicitly constraining the flatness is unnecessary, and enforcing a minimum threshold on $\hat{p}$ as a prerequisite for HCC is sufficient to guarantee the properties in Eqs. (7)-(8). In our experiments, we adopt a threshold of $0.5$, and candidate regions with lower confidence scores $\hat{p}$ are simply discarded in the pseudo labeling process. Note that enforcing the score distributions of $\mathbf{z}_{\text{sub}}$ and $\mathbf{z}_{\text{sup}}$ can be performed by simply applying a multiplicative temperature parameter in Eq. (3) set to the value in a range of $(0, 1)$.

### S1.3. LoCLIP training

**Dataset.** We construct a dedicated dataset for training LoCLIP by extracting candidate regions from an RPN [26]. Each candidate region is assigned a binary label based on its intersection over union (IoU) with any object instance in the ground-truth annotations of the base classes. Specifically, candidate regions with an IoU score greater than $0.8$ are labeled as positive, while those with an IoU score of $0.8$ or lower are labeled as negative. This process yields a dataset of image patches (*i.e.*, cropped candidate regions) paired with their binary labels. Notably, we empirically observe that LoCLIP converges quickly, and using only $1\%$ of the training split from the COCO [19] or LVIS [8] is sufficient to achieve meaningful predictions. Constructing this dataset thus requires only a negligible computational overhead.

**Loss.** The candidate regions in our curated dataset are typically imbalanced, with background regions substantially outnumbering foreground regions. Directly training LoCLIP on this imbalanced dataset tends to induce a strong bias toward the background class, leading to objectness scores that are frequently close to the value of $0$. To mitigate this issue, we adopt a weighted binary cross-entropy loss, using the ratios of foreground and background regions as weights for the positive and negative terms, respectively. This helps LoCLIP training by suppressing bias toward the background class in its predictions.

**Training.** We train LoCLIP for approximately 8k iterations using a cosine learning rate scheduler, with the maximum learning rate set to $0.001$. The parameters are optimized with SGD using a batch size of $192$ on a single NVIDIA A6000 GPU. We note that, unlike previous works [31, 32, 37, 38] which estimate the objectness of candidate regions directly from the RPN, our approach requires an additional training step for LoCLIP. However, this extra step takes no more than $5$ minutes, which is negligible in practice, while providing significantly more reliable objectness estimations for unseen object classes (Table 4).

**Hyperparameter $\tau$.** We select $\tau$ by evaluating the binary classification accuracy on the dataset used for training LoCLIP. We perform a grid search for $\tau$ in the range $[0.1, 0.9]$ with a $0.1$ interval, and choose the value that yields the best accuracy. While a more extensive approach to searching for the optimal $\tau$ could be used, *e.g.*, by curating a dedicated validation set, we found this simple method to be sufficient.

### S1.4. LLM hierarchy refinement

We establish an initial LLM-generated hierarchy by querying LLMs to provide $K$ super- and sub-categories of a given object class [20, 24]. Directly exploiting LLM responses, however, can introduce incorrect entries into the generated hierarchical structure, as LLMs are prone to make mistakes. We point out that previous methods [20, 24] that use LLMs for establishing a hierarchy structure between object classes, have typically been limited to using only a small number of $K$ for establishing the hierarchy, *e.g.*, 3 in [20]. On the one hand, using a larger $K$ allows to consider the various super-/sub-categories and helps to capture diverse semantic relationships. On the other hand, it could introduce incorrect concepts in the LLM-generated hierarchy, since the risk of LLM providing incorrect response increases as $K$ grows. To address this problem, we additionally refine the initial super-/sub-categories through an LLM-driven process based on three main criteria: 1) correctness, 2) discriminability, and 3) near-duplicate removal. We describe in the following each step in detail. Note that applying these processes may result in a varying number of super- and sub-categories per class and we have assumed in Eqs. (3) and (4) a uniform number per class for notational simplicity.

**Correctness.** For each entry given by the LLM, we ask the LLM if the given entry is indeed correct, *i.e.*, if the given association (*e.g.*, super- or sub-ordinate relationships) holds, using the "Is-A" relationship, which is the foundation of the WordNet database [23]. For example, if a *golden retriever* is given as a sub-category of *dog*, we ask the LLM: Is *golden retriever* a *dog*? Similarly, when *companion animal* is given as a super-category of *dog*, we ask the LLM: Is *dog* a *companion animal*? Although exploiting the same model

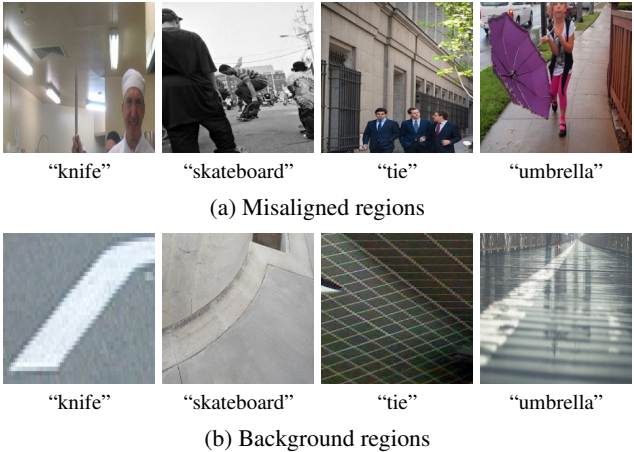

| "knife" | "skateboard" | "tie" | "umbrella" |

(a) Misaligned regions

| "knife" | "skateboard" | "tie" | "umbrella" |

(b) Background regions

Figure S1. Examples of candidate regions for novel classes in COCO [19]. We show regions whose confidence scores predicted using CLIP [25] exceeds 0.9, together with the estimated class labels.

for both generation and verification may seem redundant, this self-validation step serves as an effective filtering mechanism for incorrect entries. This observation coincides with the self-consistency strategy [33], which suggests that LLMs are capable of refining their own outputs to achieve better performance.

**Discriminability.** While many entries in the LLM-generated hierarchy could provide useful cues for discriminating objects of different classes, some might negatively affect the process by being distractive. This is particularly the case for super-categories, since many distinct objects can share a common, overly broad super-category, making it uninformative. For example, the concept *entity* is a valid super-category shared across all classes, but it provides no useful discriminatory cues. To alleviate this, we discard super-category entries that are shared by a disproportionately large number of object classes, e.g., more than $1/3$ of the total number of novel object classes on OV-COCO [19].

**Near-duplicate removal.** Lastly, we remove near-duplicates. We compute the cosine similarity between entries in the CLIP embedding space and discard one entry from any pair found to be too close. This process removes, e.g., simple singular/plural variations of the same concept, such as *home appliance* and *home appliances*.

We visualize in Fig. S4 our super- and sub-categories of novel object classes in OV-COCO [19] obtained using GPT-OSS-120b [1] as a LLM where $K$ is initially set to 30 and 10 for sub- and super-categories, respectively.

## S2. More discussions

**Limitations of CLIP confidence score.** Standard VLMs employed in OVD, such as CLIP [25] and ALBEF [15], are pre-trained for image-level tasks, leading to a discrepancy when applied to region-level predictions in OVD. Consequently, their confidence scores do not reliably reflect the presence of novel object classes in candidate regions. To demonstrate this limitation, we show in Fig. S1 the candidate regions whose confidence scores $\hat{p}$ exceeds 0.9, but correspond to misaligned (Fig. S1 (a)) or correspond to the background (Fig. S1 (b)). We can see that directly using the confidence scores $\hat{p}$ in the context of pseudo labeling is suboptimal, resulting in the generated pseudo labels to be dominated by irrelevant regions [31].

**Is confidence adjustment in HCC necessary?** With the class-wise sub-category scores $\mathbf{z}_{\text{sub}}$ and super-category scores $\mathbf{z}_{\text{sup}}$ in hand, a direct formulation of hierarchical consistency involves verifying whether the class estimations across all levels are consistent. Specifically, we can assign a pseudo class label $\hat{y}_b$ for the region $b$ as follows:

$$\hat{y}_b = \arg\max(\mathbf{p}) \text{ if }$$
$$\arg\max(\mathbf{p}) = \arg\max(\mathbf{z}_{\text{sub}}) = \arg\max(\mathbf{z}_{\text{sup}}). \tag{S10}$$

This strategy, however, is sub-optimal as it relies solely on the top-ranking indices (i.e., $\arg\max$), thereby discarding the rich statistical information (e.g., likelihood values) encoded in the full distributions of $\mathbf{z}_{\text{sub}}$, $\mathbf{z}_{\text{sup}}$, and $\mathbf{p}$. By contrast, HCC exploits these full distributions to calibrate the confidence scores, allowing for a more robust and fine-grained estimation of object presence. Quantitatively, the strategy defined in Eq. (S10) achieves 35.5 $\text{AP}_{50}^N$, which improves upon the baseline (32.2 $\text{AP}_{50}^N$) but is significantly outperformed by our HCC technique (37.8 $\text{AP}_{50}^N$), confirming the efficacy of our calibration approach.

**Reproducibility of MarvelOVD [31].** In Sec. 4.2, we report the performance of MarvelOVD [31] using our re-implementation. Note that although the official code is publicly available[1], we found that it fails to execute due to a fundamental error. We attempted to contact the authors regarding this issue but did not receive a response. Therefore, the results reported in Sec. 4.2 are obtained exclusively from our re-implementation for MarvelOVD [31], which is based on the official code provided by the authors.

## S3. More results

**More results on hierarchical consistency.** We provide additional visualizations of hierarchical consistency in

---
[1] https://github.com/wkfdb/MarvelOVD

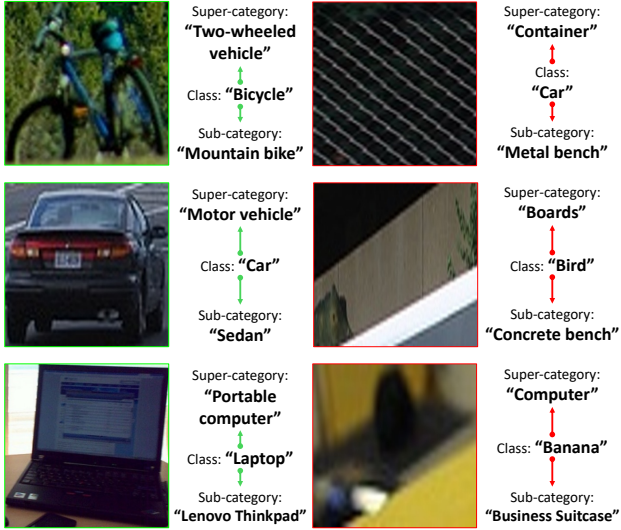

Figure S2. Visualization of hierarchical consistency of candidate regions. We group the candidate regions into well-localized (left) and irrelevant (right) regions based on their IoU with the ground-truth bounding boxes for base object classes. Green and red arrows indicate consistent and inconsistent hierarchical associations, respectively. See text for details.

Fig. S2. We can see that well-localized regions tend to yield hierarchically consistent predictions, whereas irrelevant ones do not, which aligns with the observations in Fig. 1. We further perform quantitative analysis and observe that $88.9\%$ of the well-localized region yield hierarchically consistent predictions, while only $11.2\%$ of the irrelevant regions do. The results demonstrate a significant gap in consistency ratios between the two groups, suggesting that hierarchical consistency could serve as a valuable cue for discriminating valid objects from the background.

**Comparison between RPN and LoCLIP.** The RPN [26], commonly used in previous works [31, 32, 37] for estimating the objectness of candidate regions, is sub-optimal due to its inherent bias towards base object classes. Consequently, it fails to provide reliable objectness scores for unseen object classes. To address this limitation, we introduce LoCLIP, which employs a parameter-efficient fine-tuning strategy to estimate objectness. We present qualitative comparisons of objectness scores from the RPN and LoCLIP in Fig. S3. The visualizations clearly demonstrate the advantages of LoCLIP over the RPN. For example, for novel object instances shown in the left column (*e.g.*, *bus*, *knife*, *dog* and *cow*), the RPN yields low objectness scores, failing to identify them as foreground. In contrast, LoCLIP successfully assigns high scores to these regions. Conversely, for the irrelevant background regions in the right column, the RPN tends to produce inappropriately high scores, whereas LoCLIP effectively suppresses them, demonstrating its advantage as a reliable objectness score metric.

Table S1. Performance comparison between models trained with pseudo labels obtained using different LLMs on OV-COCO [19].

| LLM | $AP_{50}^N$ | $AP_{50}^B$ |
| --- | --- | --- |
| - | 32.2 | 58.3 |
| LLaMa-3-8B [6] | 36.2 | 59.2 |
| Qwen-3-30b [30] | <u>37.4</u> | 59.4 |
| GPT-OSS-120b [1] | **37.8** | 59.4 |

**Using different LLMs.** We primarily use GPT-OSS-120b [1] as a LLM throughout all experiments in the main paper. To further demonstrate the generalization capability of our approach, we employ various LLMs and compare the results. Specifically, we establish hierarchical structure of novel object classes by extracting super-/sub-categories using LLaMa-3.1-8B [6] and Gemma-3-27b [30]. We present the results in Table S1. We can see that all entries in Table S1 outperform a baseline that does not exploit the HCC technique, demonstrating that our work can work well with various LLMs. We point out that while it is possible to construct accurate hierarchies for each dataset manually, doing so requires considerable domain expertise and effort, and such hierarchies would need to be rebuilt for every dataset in use. In contrast, LLMs provide an efficient, scalable way to automatically generate such hierarchies, significantly reducing human-driven efforts. In our work, we use LLMs for this purpose, not as a core contribution of our method, but as a practical means to obtain a reasonable hierarchy that our framework can exploit to improve pseudo label generation for OVD. Our framework assumes a reasonable hierarchical structure, rather than a perfect one. This aligns with results reported in prior works, despite the imperfect hierarchical structure, have successfully used LLM-generated hierarchies to improve performance in image classification [24] and object detection [20].

**Comprehensive comparison with the state-of-the-art methods.** Tables S2 and S3 present comprehensive comparisons with state-of-the-art methods on OV-COCO [19] and OV-LVIS [8], respectively. For completeness, we include methods that leverage auxiliary sources of supervision and specify additional datasets, and pretrained models used. The results demonstrate that our method outperforms other approaches that do not utilize additional datasets. Ours even outperforms previous works that employ DETR [42] as a detector [16, 36]. Generally, methods exploiting external datasets and pretrained models (*e.g.*, MViT [22] as a proposal generator) outperform ones that do not. We note that in the OVD setting, it is standard practice to

distinguish between these two paradigms, as they operate under fundamentally different assumptions; thus, a direct comparison would be unfair. In this context, exploring how to incorporate such additional datasets into our framework remains an interesting direction for future work.

**Visualization of pseudo labels.** We visualize in Fig. S5 the pseudo labels generated with and without the HCC and LoCLIP components. Compared to the baseline in the leftmost column, we observe that HCC discards irrelevant regions (*e.g.*, redundant boxes labelled as *dog* and incorrect class assignments as *cat* in the top row), while LoCLIP filters out severely misaligned regions (*e.g.*, an excessively large box labelled as *skateboard* in the top row). Similar observations can be made for the *cow* object in the second row. Overall, these two components complement each other, effectively preventing irrelevant regions from being assigned pseudo labels. We also present a failure case in the last row, where the tail region of an airplane is mislabelled as an *airplane*; neither HCC or LoCLIP was able to discard this noisy label. Nonetheless, our method still yields significantly less noisy pseudo labels compared to the baseline.

## S4. Limitations and future work

Although our approach is effective and sets a new state of the art, it shares limitations in other OVD methods using pseudo labels [31, 32, 37, 38]. First, we use class names of novel object classes to obtain accurate class label predictions from VLMs, which may not be available in real-world scenarios. Second, particularly on the OV-LVIS dataset [8], our method is outperformed by OVD methods exploiting additional source of supervision [2, 4, 11, 17, 18, 21, 41], *e.g.*, images with corresponding class labels [27] or captions [3, 28]. Exploring how to incorporate these additional datasets into a pseudo labeling framework would be an interesting direction for future work.

RPN: 0.598 | LoCLIP: 0.973 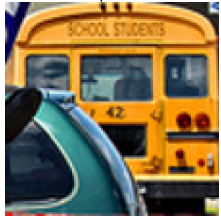 RPN: 0.574 | LoCLIP: 0.219 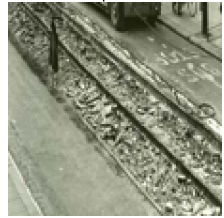

RPN: 0.284 | LoCLIP: 0.913 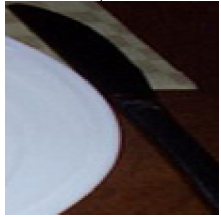 RPN: 0.564 | LoCLIP: 0.289 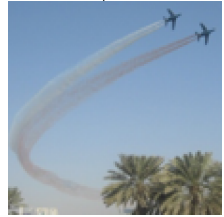

RPN: 0.030 | LoCLIP: 0.958 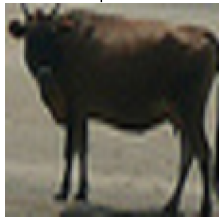 RPN: 0.794 | LoCLIP: 0.436 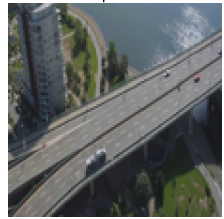

RPN: 0.381 | LoCLIP: 0.973 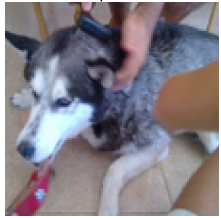 RPN: 0.790 | LoCLIP: 0.240 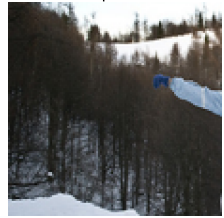

RPN: 0.650 | LoCLIP: 0.974 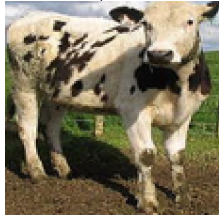 RPN: 0.745 | LoCLIP: 0.437 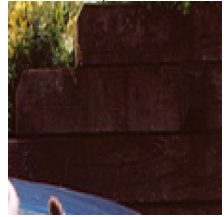

Figure S3. Qualitative comparison of objectness scores estimated by the RPN [26] and LoCLIP. LoCLIP consistently yields higher scores for valid objects and lower scores for background regions compared to the RPN, demonstrating more reliable and unbiased objectness estimations. See text for details.

(a) Super-categories for novel object classes on OV-COCO [19]

(b) Sub-categories for novel object classes on OV-COCO [19]

Figure S4. Super-/sub-categories for novel object classes in OV-COCO [19]. We establish a hierarchical structure between object classes by leveraging GPT-OSS-120b [1] as a LLM. Best viewed by zooming in.

Table S2. Comprehensive comparison with state-of-the-art methods on the OV-COCO dataset [19]. We report the mean and standard deviation across 3 independent runs. † indicates results from our re-implementation.

| Method | Backbone | Detector Architecture | Additional Supervision | Additional Pretrained Models | $AP_{50}^N$ | $AP_{50}^B$ |
|---|---|---|---|---|---|---|
| ViLD [7] | RN50 [9] | FR-CNN [26] | | | 27.6 | 59.5 |
| F-VLM [14] | RN50 [9] | FR-CNN [26] | | | 28.0 | 40.2 |
| OV-DETR [36] | RN50 [9] | DETR [42] | | | 29.4 | 52.7 |
| OADP[1] [32] | RN50 [9] | FR-CNN [26] | | | 31.3 | - |
| VL-PLM [37] | RN50 [9] | FR-CNN [26] | | | 32.3 | 54.0 |
| DK-DETR [16] | RN50 [9] | DETR [42] | | | 32.3 | 61.1 |
| RALF [13] | RN50 [9] | FR-CNN [26] | | | 33.4 | 54.5 |
| BARON [34] | RN50 [9] | FR-CNN [26] | | | 34.0 | 60.4 |
| MarvelOVD† [31] | RN50 [9] | FR-CNN [26] | | | 35.4 | 56.5 |
| SAS-Det [38] | RN50 [9] | FR-CNN [26] | | RegionCLIP [39] | 37.4 | 58.5 |
| Ours | RN50 [9] | FR-CNN [26] | | | **38.9**±0.3 | 59.5±0.2 |
| Detic [41] | RN50 [9] | FR-CNN [26] | Captions [3] | | 27.8 | 47.1 |
| DV-Det [12] | RN50 [9] | FR-CNN [26] | Captions [3] | | 35.8 | 57.0 |
| OC-OVD [2] | RN50 [9] | FR-CNN [26] | Captions [3] | MViT [22] | 36.6 | 54.0 |
| RALF [13] | RN50 [9] | FR-CNN [26] | Captions [3] | MViT [22] | 41.3 | 54.3 |
| CLIFF [17] | RN50 [9] | FR-CNN [26] | Captions [3] | MViT [22] | 41.3 | 54.1 |
| BARON [34] | RN50 [9] | FR-CNN [26] | Captions [3] | MViT [22] | **42.7** | 54.9 |
| CLIPSelf† [35] | ViT-L/14 [29] | F-ViT [35] | | | 41.3 | 65.5 |
| Ours | ViT-L/14 [29] | F-ViT [35] | | | **44.0**±0.2 | 65.8±0.1 |

Table S3. Comprehensive comparison with state-of-the-art methods on the OV-LVIS dataset [19]. We report the mean and standard deviation across 3 independent runs. † indicates results from our re-implementation.

| Method | Backbone | Detector Architecture | Additional Supervision | Additional Pretrained Models | $AP_m^N$ | $AP_m^{All}$ |
|---|---|---|---|---|---|---|
| OV-DETR [36] | RN50 [9] | DETR [42] | | | 17.4 | 26.6 |
| F-VLM [14] | RN50 [9] | Mask R-CNN [10] | | | 18.6 | 24.2 |
| BARON [34] | RN50 [9] | Mask R-CNN [10] | | | 19.2 | 26.5 |
| OADP[1] [32] | RN50 [9] | Mask R-CNN [10] | | | 19.9 | - |
| DK-DETR [16] | RN50 [9] | DETR [42] | | | 20.5 | 30.0 |
| Ours | RN50 [9] | Mask R-CNN [10] | | | **21.7**±0.4 | 26.0±0.2 |
| VL-Det [18] | RN50 [9] | CenterNet2 [40] | CC3M [28] | | 21.7 | 30.1 |
| DV-Det [12] | RN50 [9] | CenterNet2 [40] | CC3M [28] | | 23.1 | 31.2 |
| CLIFF [17] | RN50 [9] | CenterNet2 [40] | IN-L [41] | MViT [22] | 24.5 | 28.2 |
| Detic [41] | RN50 [9] | CenterNet2 [40] | IN-L [41] | | 24.6 | 32.4 |
| OC-OVD [2] | RN50 [9] | CenterNet2 [40] | IN-L [41] | MViT [22] | **25.2** | 32.9 |
| CLIPSelf† [35] | ViT-B/16 [29] | F-ViT [35] | | | 25.1 | 24.5 |
| Ours | ViT-B/16 [29] | F-ViT [35] | | | **25.5**±0.2 | 24.7±0.1 |

[1] The results for OADP [32], reported in the original publication, were obtained using LSJ data augmentation [5], which is inconsistent with the experimental setting described in the paper. We thus report the results without the LSJ technique, obtained from the official repository.

(a) HCC ✗,  LoCLIP ✗   (b) HCC ✓,  LoCLIP ✗   (c) HCC ✗,  LoCLIP ✓   (d) HCC ✓,  LoCLIP ✓

Figure S5. Visualizations of pseudo labels on OV-COCO [19] with and without the HCC and LoCLIP components. Green and orange boxes represent base and novel object classes, respectively. The usage of each component is marked by ✓ and ✗ . The bottom row shows a failure case. Best viewed in color.




\end{document}